\pdfoutput=1

\documentclass[11pt]{article}

\usepackage[final]{acl}

\usepackage{times}
\usepackage{latexsym}

\usepackage[T1]{fontenc}

\usepackage[utf8]{inputenc}

\usepackage{microtype}

\usepackage{inconsolata}

\usepackage{amsmath,amsfonts}
\usepackage{algorithmic}
\usepackage{graphicx}
\usepackage{textcomp}
\usepackage{xcolor}
\usepackage{algorithm}
\usepackage{multirow}
\usepackage{booktabs}  
\usepackage{bm}
\usepackage{arydshln}

\usepackage{hyperref}
\usepackage{url}

\usepackage[inkscapelatex=false]{svg}
\usepackage{amsmath}
\usepackage{amssymb}
\usepackage{colortbl}
\usepackage{arydshln}
\usepackage{booktabs}
\usepackage{multirow}
\usepackage{enumerate}
\usepackage{makecell}
\usepackage{tabularx}
\usepackage{threeparttable}
\usepackage{picture}
\usepackage{verbatim}
\usepackage{algorithm}
\usepackage{algorithmic}
\usepackage{graphicx}
\usepackage{subcaption}

\title{PACE: Improving Prompt with Actor-Critic Editing for Large \mbox{Language Model}}

\author{\textbf{Yihong Dong},
    \textbf{Kangcheng Luo},
    \textbf{Xue Jiang},
  \textbf{Zhi Jin}, 
   and \textbf{Ge Li}\\
   Key Laboratory of High Confidence Software Technologies (Peking University), \\ Ministry of Education; School of Computer Science, Peking University, Beijing, China \\
    \texttt{\{dongyh, luokangcheng, jiangxue\}@stu.pku.edu.cn}, 
    \texttt{\{zhijin, lige\}@pku.edu.cn} \\ 
}

\begin{document}

\maketitle

\begin{abstract}
Large language models (LLMs) have showcased remarkable potential across various tasks by conditioning on prompts. However, the quality of different human-written prompts leads to substantial discrepancies in LLMs' performance, and improving prompts usually necessitates considerable human effort and expertise. To this end, this paper proposes Prompt with Actor-Critic Editing (PACE) for LLMs to enable automatic prompt editing. Drawing inspiration from the actor-critic algorithm in reinforcement learning, PACE leverages LLMs as the dual roles of actors and critics, conceptualizing prompt as a type of policy. PACE refines prompt, taking into account the feedback from both actors performing prompt and critics criticizing response. This process helps LLMs better align prompt to a specific task, thanks to real responses and thinking from LLMs.
We conduct extensive experiments on 24 instruction induction tasks and 21 big-bench tasks. Experimental results indicate that PACE elevates the relative performance of medium/low-quality human-written prompts by up to 98\%, which has comparable performance to high-quality human-written prompts. Moreover, PACE also exhibits notable efficacy for prompt generation. 
\end{abstract}

\section{Introduction}
The rapid development of LLMs has led to notable advancements in artificial intelligence. LLMs, such as ChatGPT \citep{ChatGPT}, have emerged as essential tools in various application scenarios, including automatic question-answering \cite{DBLP:conf/acl/GabburoGKM23, DBLP:conf/nips/CartaOSL22}, embodied agent \cite{DBLP:conf/aaai/LanchantinSSSSS23, DBLP:conf/aaai/Seraj23}, and code generation \citep{Subtoken-TranX, Self-planning, CODEP}, among others. They have demonstrated remarkable capabilities in handling a range of tasks. However, their efficacy is not universal and often depends on how we interact with them - namely, how we provide appropriate prompts. 

\begin{figure}[t!]
    \centering
    \includegraphics[width=0.53\textwidth]{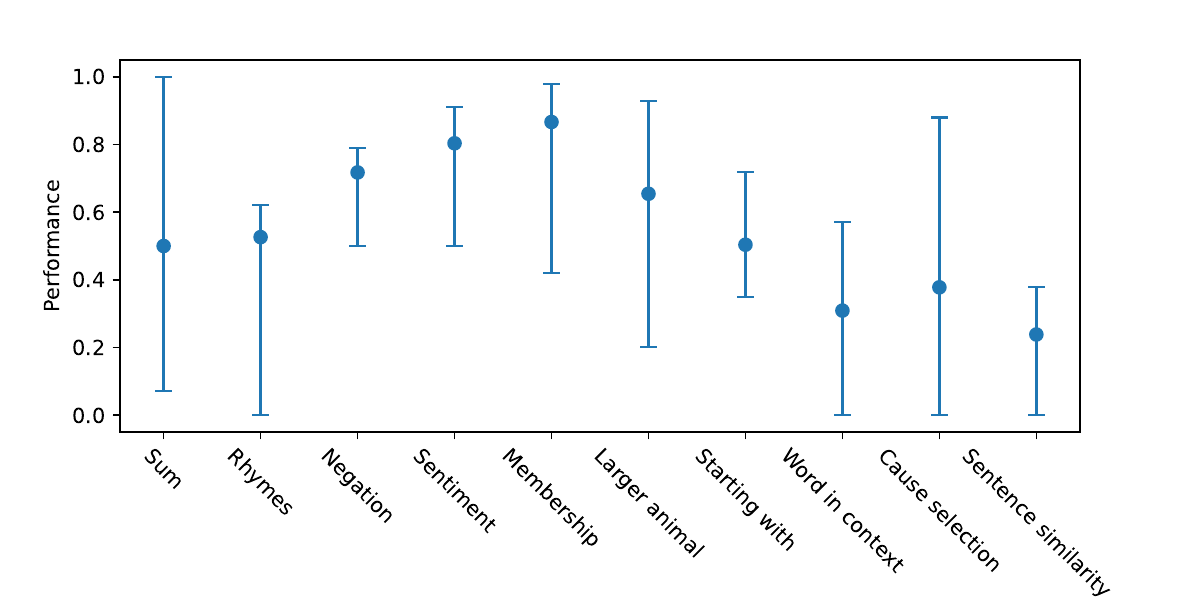}
    \caption{The human-written prompt performance of ten tasks proposed in Instruction Induction dataset \citep{Instruction_Induction}, where each task contains about eight human-written prompts, with absolute performance differences between 29\% and 93\% for each task (refer to Appendix \ref{DII} for detailed results).}
    \label{fig: Instruction_Induction}
\end{figure}

The interaction between users and LLMs is heavily mediated by prompts. These prompts can be understood as entry points, shaping and directing the LLMs' enormous reserves of knowledge and computational abilities toward specific outcomes. Therefore, just as a key unlocks a door or a command instructs a computer program, prompts guide the response mechanisms of LLMs, determining the range and depth of answers they provide. Figure \ref{fig: Instruction_Induction} vividly illustrates the sensitivity of LLMs to the quality and specificity of prompts. Even when posed with the same underlying task, varying the phrasing or approach of a prompt can yield vastly different results. For instance, while one prompt might retrieve a broad overview, another, only slightly rephrased (adding ``in detail''), could elicit a detailed response. This variability underscores not only the importance of crafting thoughtful and effective prompts but also the nuanced complexities embedded within this task.

The variability in LLM performance with prompts is primarily caused by two key factors: 1. Human Articulation Limitations: Humans inherently think and communicate in a manner that is filled with nuances, emotions, and subjectivities. When posing questions or presenting requirements, they might inadvertently omit vital details or introduce ambiguities. Our natural way of communication is also peppered with cultural references, idioms, and shorthand that might not always be clear or universally understood. Therefore, human-written prompts can sometimes express incomplete, ambiguous, or even erroneous requirements. This not only affects the accuracy of the output but might also skew it in unintended directions. 2. Cognitive Discordance between Humans and LLMs: Even when we assume that a prompt is perfectly articulated, another challenge arises. There is an intrinsic cognitive gap between human comprehension and the way language models interpret information. Human comprehension and cognition need to be ``translated'' into expressions that align with LLM's expectations. Thus, a well-phrased requirement from a human perspective may still lead to an LLM output that feels off or unexpected to humans. As a result, crafting high-quality prompts is usually not accomplished in one go but requires trial and error.

In general, humans develop prompts in two stages: 1. Summarizing the initial prompt: The first stage involves crafting a preliminary version of a prompt. It is a process that often draws upon existing data, prior knowledge, or a specific need within a given context. 2. Improving and editing prompts: The second stage is characterized by a continuous process of improvement, modification, and fine-tuning. This stage is crucial because it takes the initial draft to a polished level, where the prompt becomes more precise, clear, and potentially more effective in eliciting the desired response from an LLM. This stage often involves iterative feedback loops, close scrutiny. However, existing approaches of automatic prompt engineering concentrate primarily on the first stage \cite{PromptProgramming, Instruction_Induction, APE}, while overlooking the second stage that significantly enhances the quality of prompts. Intuitively, the effect and human effort of the second stage far surpass those of the first stage, as it is relatively straightforward for humans to draft a preliminary version of the prompt. Consequently, there emerges an imperative need for advancements in automatic prompt editing for LLMs.

In this paper, we propose an effective approach named PACE to tackle automatic prompt editing for LLMs. This approach draws inspiration from the actor-critic algorithm, a well-known technique in the realm of reinforcement learning. PACE repurposes LLMs as both the actor and the critic. Conceptually, a prompt can be viewed as a policy that directs the behavior of the LLM. The actor, or the LLM, performs tasks based on this policy (prompt), while the critic provides a form of supervision, identifying how well the actor is performing based on the provided prompts. On this basis, PACE improves the quality of the prompt, thereby optimizing the performance of LLMs on specific tasks. We conduct extensive experiments to evaluate the effectiveness of PACE on 24 instruction induction tasks and 21 big-bench tasks. The experimental results indicate the effectiveness of PACE for automatic prompt editing and generation. 

\begin{figure*}[th!]
    \centering    
    \includegraphics[width=0.98\textwidth]{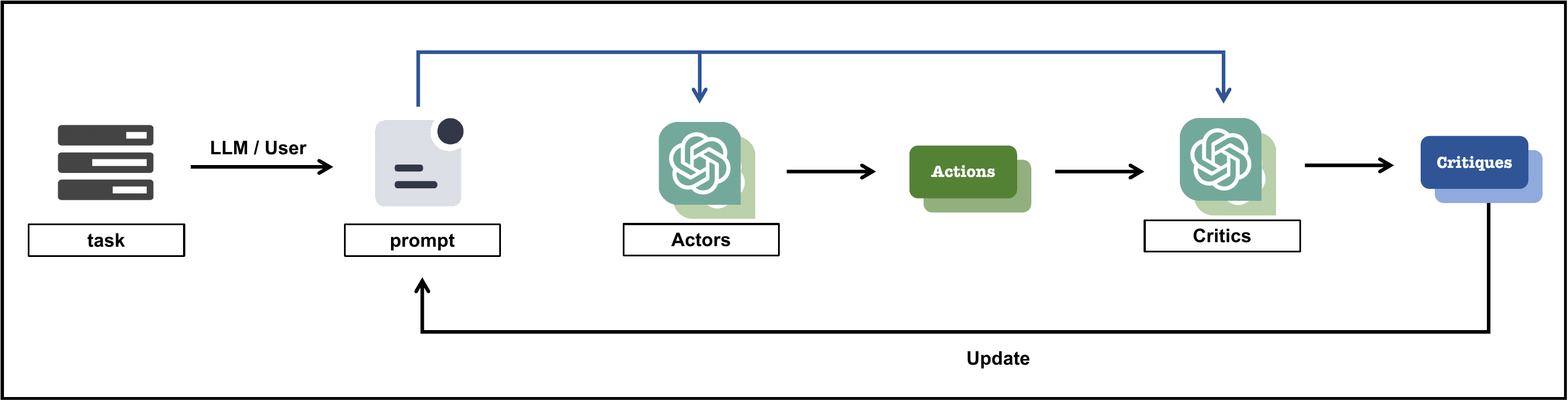}
    \caption{The paradigm of PACE.}
    \label{fig: paradigm}
\end{figure*}

\section{PACE}
We consider a task $\mathcal{T}$, accompanied by demonstration data \( D = \{(X, Y)\} \). For the task $\mathcal{T}$, given an input \( X \), a corresponding desired output is \( Y \). The objective of automatic prompt editing is to identify a prompt \( p \) such that, when a LLM $\mathcal{M}$ is queried with the concatenation of this prompt and a specified input \([p; X]\), it generates the corresponding output \( Y \). Therefore, we aim to find the prompt \( p \) that maximizes the expectation of score \( s(p, X, Y) \) over possible pairs \( (X, Y) \). 

\begin{equation*}\label{eq:score}
p^{\star} = \arg\max_p s(p) =  \arg\max_p \mathbb{E}_{(X, Y)}{s(p, X, Y)},
\end{equation*}
where $\arg\max$ means to return the parameter that is the maximum value of the function, $\mathbb{E}$ indicates the expectation, and $s$ is a score function. 

Owing to the vast and potentially infinite search space, there are significant challenges in obtaining optimal prompts. Following the previous work \cite{PromptProgramming, Instruction_Induction, APE}, we leverage the capabilities of LLM to approximate the inference of the most likely prompt $p$ with a high score. To further augment the proficiency of LLM in generating or refining prompts, we introduce the PACE approach, which consists of the actor-critic paradigm and iterative algorithm.

\subsection{Actor-Critic Paradigm}
As shown in Figure \ref{fig: paradigm}, given a prompt generated by LLM or human, we use LLMs in dual roles: as both the actor and the critic. Prompt $p$ in this context is conceptualized as a policy guiding the LLM. The better the prompt, the more effective the LLM's response to a particular task. A policy, in reinforcement learning terms, is a strategy that the model uses to determine its actions based on its current state. By treating prompts as a policy, we can leverage the concepts of reinforcement learning to guide the iterative refinement of the prompts. Specifically, in PACE, we use role instruction \cite{Self-collaboration} to construct both actors and critics:

\textbf{Actor} refers to the LLM that executes a task based on a given prompt. The response generated by the LLM is a direct consequence of the prompt, and it serves as an action taken by the actor in the environment of natural language processing tasks. Formally, for a prompt \( p \) and an input \( X \), the action \( a \) generated by the actor can be represented as:
\begin{equation}
    a = f_{\operatorname{actor}}([p; X], \mathcal{M}),
    \label{a}
\end{equation}
where $f_{*}$ is regarded as the mapping of LLM from input to output. Specifically, the input parameter in [ ] will be represented in the form of the corresponding template\footnote{The template can be found in Appendix \ref{Prompt}.}, which is then fed into the given LLM $\mathcal{M}$ to obtain the output. 

\textbf{Critic} refers to the LLM that evaluates the effectiveness of the response generated by the actor. Specifically, the critic will assess if the response effectively addresses the task defined by the prompt. The feedback from the critic is then used to adjust and optimize the prompts, allowing for a continuous cycle of prompt improvement. For the response $a$, the critic then evaluates $[p; X; a]$\footnote{PACE is not employed during the testing phase, i.e., Eq. \eqref{c} does not use test data.} and the desired output $Y$ to derive a critique \( c \) as:
\begin{equation}
    c = f_{\operatorname{critic}}([p; X; a; Y], \mathcal{M}).
    \label{c}
\end{equation}

\subsection{Iterative Algorithm}
The goal of PACE is to refine the prompt to optimize the LLM's performance for a specific task. The process involves iteratively improving the prompt using feedback from the actor and the critic.

Considering the bias caused by inputs and sampling can render critiques imprecise, thus affecting the outcomes of prompt editing. In a single iteration, the PACE approach employs $n$ actors to execute the given prompt across varied inputs. Concurrently, $n$ critics evaluate the performance of these actors, providing constructive criticism. This process culminates in the aggregation of $n$ feedback opinions, offering more holistic guidance for prompt editing. Note that $n$ is a hyperparameter and is set to 4 in this paper. 

We start with an initial prompt \( p_0 \), where $p_0$ can be an empty prompt, which is equivalent to generating from scratch based on LLM. For each iterative step \( t \), the candidate prompt $p_t$ can be refined according to the aggregation of $n$ feedback $c_{\leq n}$ as:

\begin{equation}
    p_{t+1} = f_{\operatorname{update}}([p_t; c_{\leq n}], \mathcal{M}).
    \label{u}
\end{equation}

To evaluate the efficacy of candidate prompt \(p_t\) in each iteration, we employ a score function \(s\) to assess \(p_t\) based on demonstration or valid data. The score function can be broadly categorized into two types:

1. \textbf{Log Probability} involves leveraging an LLM to compute the log probability of the output \(Y\). Intuitively, a prompt that can produce an answer with a higher log probability is more likely to be selected in practical applications. 

2. \textbf{Practical Evaluation Metric} entails generating samples directly and then assessing them using the practical evaluation metric of the task, such as Accuracy, BLEU \cite{Bleu}, BertScore \cite{BERTScore}, and so forth. 

In this paper, we focus on the second type of score function, for two primary reasons: firstly, some LLMs, owing to competitive business considerations, might not disclose generation probabilities; secondly, employing the practical evaluation metric tends to bridge the disparity during testing, generally resulting in enhanced performance.

We continue the iteration until convergence is achieved or until the predefined maximum number of iterations is reached. The prompt $p^{\star}$ derived from this iterative process serves as the finalized prompt tailored for the specific task. The detailed pseudocode of PACE is outlined in Algorithm \ref{algorithm1}.

\begin{algorithm}[h!]               
\caption{Pseudocode of PACE approach.}\label{algorithm1}
\small{
\begin{algorithmic}[1]
\REQUIRE{Initial prompt $p_0$, Demonstrate data $D$ of Task $\mathcal{T}$, Score function $s$, and LLM $\mathcal{M}$.}
\ENSURE{Prompt $p^{\star}$.}\\
\STATE Initial $t=0$ and $p^{\star}=p_0$.
\REPEAT
    \FOR{i from 1 to $n$}
        \item  Sample $(X_i,Y_i)$ from $D$.
        \item  $i$-th actor $A_i$ generates an action \( a_i \) via Eq. \eqref{a}. 
        \item  $i$-th critic $C_i$ evaluates \( a_i \) to yield a critique $c_i$ via Eq. \eqref{c}. 
    \ENDFOR
    \STATE $p_t$ is updated base on $c_{\leq n}$ via Eq. \eqref{u}. 
    \STATE $p^{\star} = max(s(p^{\star}), s(p^{t+1}))$.
    \STATE $t = t + 1$.
\UNTIL{Convergence or a maximum number of iterations.}
\RETURN{$p^{\star}$}
\end{algorithmic}}
\end{algorithm}

\begin{table*}[ht!]
\centering
\caption{The Performance of PACE under Various Initial Prompts on Instruction Induction}
\label{tab:my-table}
\resizebox{0.95\textwidth}{!}{
\begin{tabular}{@{}lrrrrrrrrr@{}}
\toprule
\textbf{Task}                      & \multicolumn{1}{c}{\textbf{W-HWP}} & \multicolumn{1}{c}{\textbf{+ PACE}} & \multicolumn{1}{c}{\textbf{M-HWP}} & \multicolumn{1}{c}{\textbf{+ PACE}} & \multicolumn{1}{c}{\textbf{B-HWP}} & \multicolumn{1}{c}{\textbf{+ PACE}} & \multicolumn{1}{c}{\textbf{Butter Fingers}} & \multicolumn{1}{c}{\textbf{+ PACE}} & \multicolumn{1}{c}{\textbf{APE}} \\ \cmidrule(r){2-3} 
 \cmidrule(r){4-5}  \cmidrule(r){6-7} \cmidrule(r){8-9} \cmidrule(r){10-10}
\textbf{active\_to\_passive}       & 1                  & 0.99              & 1                  & 1                  & 1                  & 1                & 0.02                                        & 1                                   & 1                                \\
\textbf{antonyms}                  & 0.77               & 0.85              & 0.82               & 0.86               & 0.85               & 0.87             & 0.76                                        & 0.81                                & 0.82                             \\
\textbf{cause\_and\_effect}        & 0                  & 0.53              & 0.36               & 0.73               & 0.84               & 0.85             & 0.04                                        & 0.61                                & 0.5                             \\
\textbf{common\_concept}           & 0.05               & 0.06              & 0.08               & 0.15               & 0.15               & 0.16             & 0.01                                        & 0.04                                & 0.04                             \\
\textbf{diff}                      & 0.87               & 1                 & 0.94               & 1                  & 1                  & 1                & 0.92                                        & 1                                   & 0.88                             \\
\textbf{first\_word\_letter}       & 0.6                & 1                 & 0.8                & 1                  & 1                  & 1                & 0                                           & 1                                   & 1                                \\
\textbf{informal\_to\_formal}      & 0.46               & 0.59              & 0.52               & 0.6                & 0.64               & 0.67             & 0.57                                        & 0.54                                & 0.5                              \\
\textbf{larger\_animal}            & 0.2                & 0.53              & 0.46               & 0.93               & 0.93               & 0.95             & 0                                           & 0.26                                & 0.49                             \\
\textbf{letters\_list}             & 0.56               & 1                 & 0.73               & 1                  & 1                  & 1                & 0.02                                        & 0.91                                & 1                                \\
\textbf{negation}                  & 0.5                & 0.76              & 0.63               & 0.82               & 0.79               & 0.83             & 0                                           & 0.78                                & 0.81                             \\
\textbf{num\_to\_verbal}           & 0.44               & 1                 & 0.59               & 1                  & 1                  & 1                & 0.27                                        & 1                                   & 0.98                             \\
\textbf{ortho\_starts\_with} & 0.35               & 0.44              & 0.36               & 0.52               & 0.72               & 0.71             & 0.47                                        & 0.37                                & 0.42                             \\
\textbf{rhymes}                    & 0                  & 0.56              & 0.56               & 0.6                & 0.61               & 0.61             & 0.3                                         & 0.57                                & 0.12                             \\
\textbf{second\_word\_letter}      & 0.95               & 1                 & 0.96               & 1                  & 0.99               & 1                & 0.31                                        & 1                                   & 0.25                             \\
\textbf{sentence\_similarity}      & 0                  & 0.42              & 0.2                & 0.28               & 0.38               & 0.35             & 0                                           & 0.01                                & 0.11                             \\
\textbf{sentiment}                 & 0.5                & 0.91              & 0.66               & 0.91               & 0.91               & 0.92             & 0                                           & 0.89                                & 0.85                             \\
\textbf{singular\_to\_plural}      & 0.99               & 1                 & 0.99               & 1                  & 1                  & 1                & 0.98                                        & 0.99                                & 1                                \\
\textbf{sum}                       & 0.07               & 1                 & 0.99               & 1                  & 1                  & 1                & 0.64                                        & 1                                   & 0.37                             \\
\textbf{synonyms}                  & 0.11               & 0.12              & 0.13               & 0.16               & 0.15               & 0.17             & 0.12                                        & 0.14                                & 0.39                             \\
\textbf{taxonomy\_animal}          & 0.42               & 0.92              & 0.74               & 0.85               & 0.98               & 0.96             & 0.28                                        & 0.86                                & 0.69                             \\
\textbf{translation\_en-de}        & 0.81               & 0.84              & 0.82               & 0.84               & 0.84               & 0.84             & 0.8                                         & 0.83                                & 0.81                             \\
\textbf{translation\_en-es}        & 0.87               & 0.83              & 0.87               & 0.88               & 0.9                & 0.89             & 0.82                                        & 0.77                                & 0.88                             \\
\textbf{translation\_en-fr}        & 0.88               & 0.88              & 0.88               & 0.86               & 0.89               & 0.88             & 0.78                                        & 0.91                                & 0.86                             \\
\textbf{word\_in\_context}         & 0                  & 0.16              & 0.28               & 0.57               & 0.54               & 0.58             & 0                                           & 0.49                                & 0.23                             \\
\textbf{Average}                   & \textbf{0.47}      & \textbf{0.72}     & \textbf{0.64}      & \textbf{0.78}      & \textbf{0.79}      & \textbf{0.8}     & \textbf{0.36}                               & \textbf{0.71}                       & \textbf{0.62}                    \\   \bottomrule                   
\end{tabular}}
\end{table*}

\section{Experiment Setup}

\paragraph{Benchmarks.} We perform a comprehensive evaluation on Instruction Induction \cite{Instruction_Induction} and Big-Bench \cite{BIG-Bench} to demonstrate the efficacy of PACE. 

\textbf{Instruction Induction} \cite{Instruction_Induction} consists of 24 diverse instruction induction tasks, each comprising a multitude of human-written prompts. These tasks cover numerous areas of language understanding, ranging from fundamental sentence structures to the identification of similarities and causal relationships. 

\textbf{Big-Bench} \cite{BIG-Bench} is composed of 21 more challenging tasks covering many aspects of language understanding, including emotional understanding, context-free questions and answers, reading comprehension, summaries, algorithms, and various reasoning tasks. Each task has a human-written prompt.

We followed the setup of APE \cite{APE}, dividing over 40 tasks in two benchmarks into train, val, and test sets. Our approach is optimized only on the train and val sets, and the final generated prompt is evaluated on the test set.
Detailed descriptions of each task in two benchmarks can be found in Appendix \ref{app:imp_details}. 

\paragraph{Implementation Details.} In all experiments, we invoke ChatGPT as our base model through its API, namely gpt-3.5-turbo. We employ `0301' version of gpt-3.5-turbo, which is a snapshot from March 1st 2023, and will not receive updates. To increase the stability of LLM's output, we set the decoding temperature to 0 and top\_p to 1. Moreover, we set max\_tokens to 512 for generation. For hyperparameters of PACE, we set the number of agents $n$ to 4 and that of candidates in each iteration to 2. For fairness, the number of candidates in other approaches is set to 4*2 = 8. Unless otherwise stated, the maximum number of iterations is set to 1, i.e., we use only 1 iteration step for prompt editing in total. The experiments are run five times and the average results are reported.

\section{Experimental Results}

In this section, we detail the results of our comprehensive experiments which offer compelling evidence of the effectiveness of PACE in improving the performance of LLMs. 
Note that in most experiments, we only present the average results from all experiments for each benchmark. Detailed results for each task are available in the Appendix. 

\subsection{The Effect of PACE in Prompt Editing}
In \textbf{Instruction Induction}, we evaluated the performance of PACE under various initial prompts, which included: 
\begin{itemize}
    \item \textbf{Worst Human-Written Prompt (W-HWP)}: The least effective prompt among all human-written prompts included in the task\footnote{We evaluate all human-written prompts of the task on the base model and then rank their performance. Detailed result can be find in Appendix \ref{DII}};
    \item \textbf{Medium Human-Written Prompt (M-HWP)}: Its efficacy is at the median compared to all human-written prompts in the task;
    \item \textbf{Best Human-Written Prompt (B-HWP)}: Out of all human-written prompts provided in the task, it yielded the best results; 
    \item \textbf{Butter Fingers}: The variant of M-HWP with a 15\% misspelling rate introduced randomly.
\end{itemize}

As shown in Table \ref{tab:my-table}, we observe that PACE is effective with human-written prompts of varying quality. PACE was successful in substantially enhancing the performance of LLMs that were initially provided with medium-quality and low-quality human-written prompts, including M-HWP, W-HWP, and Butter Fingers. In many cases, the LLMs using the PACE-refined prompts achieved performance levels comparable to, and in some cases even surpassing, those using high-quality human-written prompts, i.e., B-HWP. Remarkably, even for B-HWP, PACE manages to offer a marginal improvement. A notable highlight is the performance of PACE under the Butter Fingers setting, which encapsulates reading comprehension challenges. Prompts under this category can be notoriously difficult, often with inherent errors or misconstructions. However, the ability of PACE to detect, correct, and improve these prompts is nothing short of commendable. A staggering enhancement of up to 98\% in the LLM's performance is a testament to PACE's robust error rectification capabilities. Equally impressive is the breadth of PACE's effectiveness. These improvements aren't isolated to specific tasks or certain domains. 
On the contrary, a consistent positive trend is observed across a diverse suite of 24 tasks, suggesting the generalizability of PACE.

It has been observed that the performance of APE \cite{APE} is comparable to that of a medium human-written prompt. However, our proposed PACE outperforms APE, even under challenging conditions like the worst human-written prompts and the "Butter Finger" settings, underscoring the superiority of our approach. We also compare the efficiency of PACE and APE and find that the running time of PACE is slightly lower than APE (about 0.78$\times$), which is acceptable.
Moreover, it's essential to highlight that for many tasks, especially those requiring an initial draft or a general directive, humans can often produce a satisfactory first attempt without much effort. For instance, humans can provide a broad overview or a general description of the intended subject. The real challenge, and where computational models like PACE come into play, is refining and optimizing these drafts to produce a high-quality final product. 
\begin{figure}[th!]
    \centering
    \begin{subfigure}{0.43\textwidth}
        \centering
        \includegraphics[width=\textwidth]{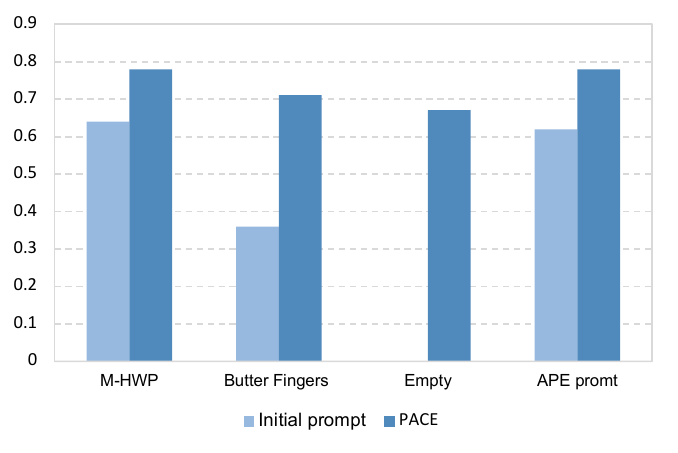}
        \caption{Instruction Induction.}
    \end{subfigure}
    \hfill
    \begin{subfigure}{0.43\textwidth}
        \centering
        \includegraphics[width=\textwidth]{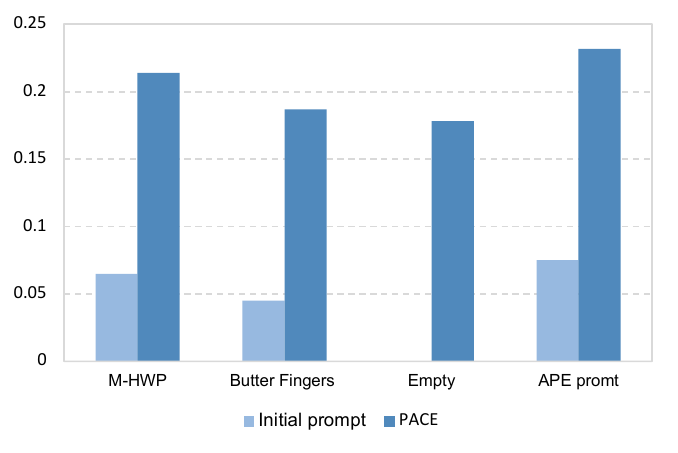}
        \caption{Big-Bench.}
    \end{subfigure}
    \caption{The Performance of PACE under Various Initial Prompts.}
    \label{fig: e12}
\end{figure}

In \textbf{Big-Bench}, each task is provided with only a single instruction, which limits our ability to screen prompts of varying qualities, unlike the tasks taken in Instruction Induction. For this reason, the only instruction we have by default is \textbf{M-HWP}, and in addition to the \textbf{Butter Fingers} setting, we have introduced two new settings: 
\begin{itemize}
    \item \textbf{Empty}: The initial prompt is an empty string; 
    \item \textbf{APE prompt}: The initial prompt is generated by LLM with APE \cite{APE}.
\end{itemize}

Figure \ref{fig: e12} elucidates the impact of PACE on both Instruction Induction and Big-Bench across four distinct settings. It is evident that PACE exhibits consistent improvements across all four settings, highlighting its robust capability to navigate through these specific conditions effectively. The enhancement with the application of PACE on the APE prompt means that even for other LLM-generated prompts, PACE can be further improved and enhanced to achieve better results, because PACE takes into account realistic feedback and LLM cognitive processes. It is worth noting the effect shown by PACE when initialized with an Empty prompt. This implies that PACE’s utility is not confined to the mere editing of pre-existing prompts. It is equally adept at crafting initial prompts from scratch, underlining its versatile and comprehensive applicative potential. The flexibility demonstrates PACE's versatility and its potential in a wide array of scenarios. Furthermore, the comparative analysis between PACE and APE reveals the superiority of PACE in enhancing performance across both two benchmarks.

In conclusion, the results from Figure \ref{fig: e12} accentuate the effectiveness and adaptability of PACE across different benchmarks and settings. Whether refining existing prompts or creating new ones, PACE consistently delivers enhanced results.

\begin{figure}[th!]
    \centering
    \begin{subfigure}{0.43\textwidth}
        \centering
        \includegraphics[width=\textwidth]{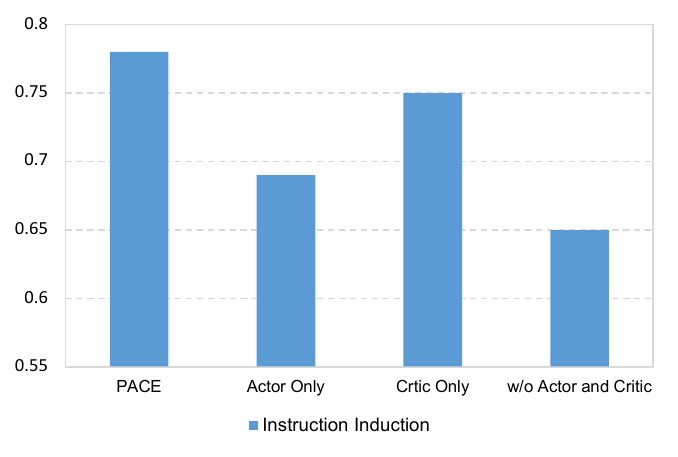}
    \end{subfigure}
    \hfill
    \begin{subfigure}{0.43\textwidth}
        \centering
        \includegraphics[width=\textwidth]{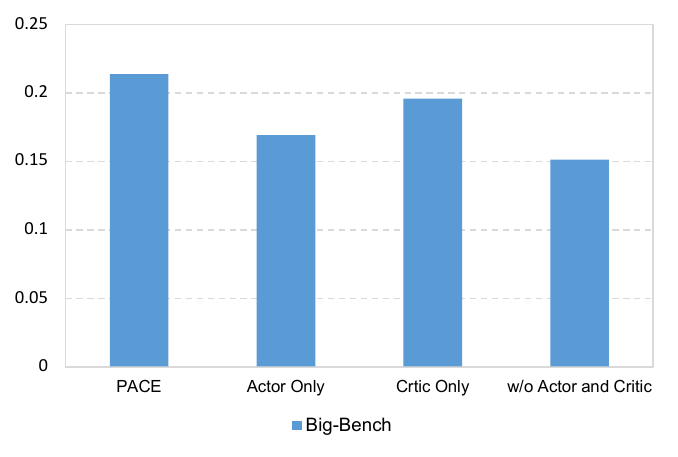}
    \end{subfigure}
    \caption{The ablation Study of PACE on Both Two Public Benchmark Datasets.}
    \label{fig: ab12}
\end{figure}

\subsection{Ablation Study}
In this section, we delve deeper into the analysis of PACE through an ablation study, which is designed to gauge the individual contributions and effectiveness of each module incorporated in APCE.

Figure \ref{fig: ab12} provides a clear visual representation of our findings. We observed that both roles – the actor and the critic – are instrumental in the overall efficiency of PACE. The actor's primary function is to execute the prompt, offering real-time feedback to the LLM. This feedback is not merely mechanical but is crucial in dynamically shaping the prompt based on changing conditions or requirements. On the other hand, the critic operates at a meta-level, assessing the quality and relevance of the feedback. Through thoughtful evaluation, the critic aids the LLM in refining and editing the prompt to ensure optimal results. 

Comparative analysis between the two roles reveals that the critic possesses a slightly higher significance in enhancing the system's performance, followed closely by the actor. The critic's evaluative capabilities ensure that the system doesn't veer off-course, while the actor provides the necessary operational feedback to keep the system in check. It is also worth noting the stark difference in performance when these roles are absent. Methods that do not incorporate the actor and critic mechanisms lag noticeably in effectiveness. This disparity is evident on both benchmarks we tested, underscoring the importance of these components in PACE.

In essence, our ablation study underscores the synergistic relationship between the actor and critic in PACE. While each has its unique function, together they substantially elevate the system's efficiency and accuracy.

\begin{figure}[th!]
    \centering
    \begin{subfigure}{0.39\textwidth}
        \centering
        \includegraphics[width=\textwidth]{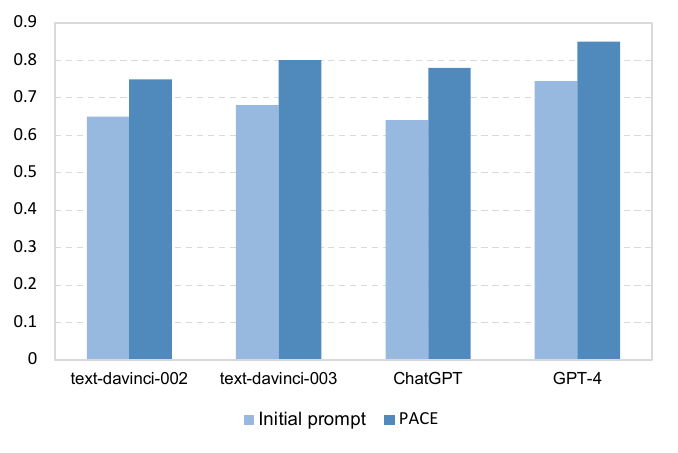}
        \caption{Instruction Induction.}
    \end{subfigure}
    \hfill
    \begin{subfigure}{0.39\textwidth}
        \centering
        \includegraphics[width=\textwidth]{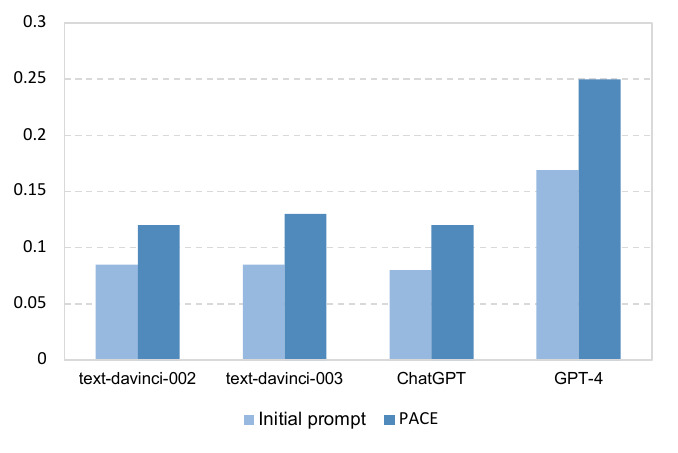}
        \caption{Big-Bench.}
    \end{subfigure}
    \caption{Performance of PACE with Different LLMs.}
    \label{fig: d12}
\end{figure}

\subsection{Comparison with different LLMs}
In this section, our aim is to underscore the versatility and generality of the PACE methodology. To do so, we have decided to utilize an array of different LLMs for the PACE task, including `text-davinci-002', `text-davinci-003', `ChatGPT', and `GPT-4'. This diverse selection not only showcases the breadth of models available but also ensures a comprehensive assessment across different model capabilities and specializations. 

Referring to Figure \ref{fig: d12}, the visual representation distinctly showcases that irrespective of the model chosen, the PACE method consistently enhances the quality of the initial prompt. This not only strengthens the argument for the efficacy of PACE but also demonstrates the robustness of the LLMs in refining textual inputs. This observation is pivotal, as it suggests that the approach is model-agnostic to some extent, and the gains are not just circumstantial or confined to specific LLMs.

In summary, the consistent improvement observed across diverse models unequivocally demonstrates that the PACE methodology serves as a universally applicable technique. This technique is instrumental in enhancing the performance of various LLMs by refining the prompts with which they are provided.

\begin{figure*}[h!]
\centering
\includegraphics[width=0.92\textwidth]{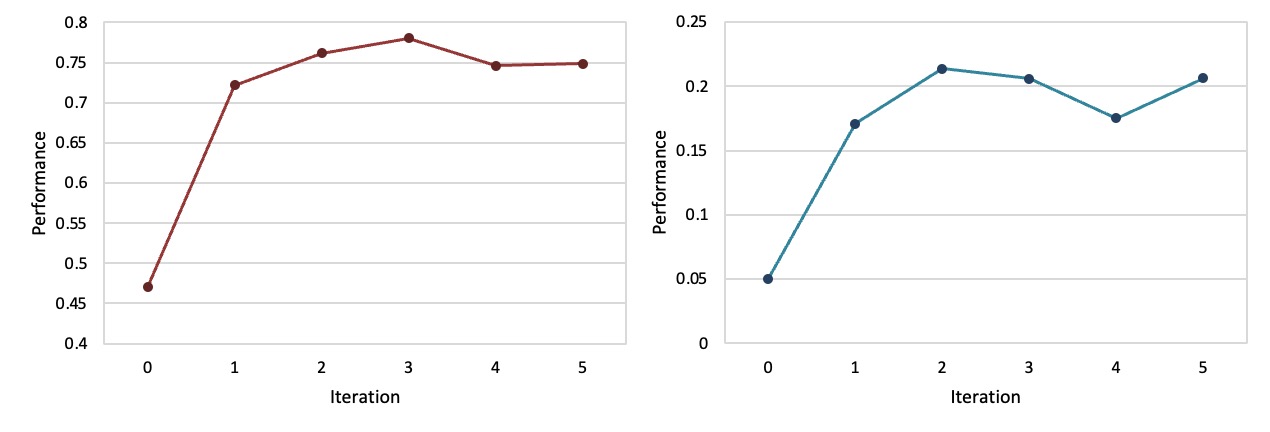}
\footnotesize{\qquad(a) Instruction Induction. \qquad \qquad  \qquad \qquad\qquad\qquad \qquad \quad (b) Big-Bench.}
\caption{Effect of Iteration numbers
}
\label{Effect of Iteration numbers}
\end{figure*}

\subsection{Effect of Iteration numbers}
In this section, our primary objective is to assess the impact of varying the number of iterations on our process or system. It's crucial to understand how iteration numbers can influence the outcomes, as this can shed light on the stability, efficiency, and effectiveness of the procedure in question. By systematically altering the iteration count, we can derive insights into the optimal number needed to achieve the desired results without overcomplicating or overburdening the system. 

In the exploration of the impact of iteration numbers, it's pivotal to understand how iterations influence the outcome. As depicted in Figure \ref{Effect of Iteration numbers}, there's a clear trend showcasing the correlation between the number of iterations and the editing effect. The pattern suggests a dynamic evolution, wherein the editing effect witnesses a surge with increasing iterations, up to a point beyond which the effect plateaus and eventually stabilizes. 

This stabilization of the editing effect after a certain number of iterations indicates a saturation point or a threshold beyond which additional iterations don't contribute significantly to enhancing the effect. What's noteworthy from the observed data is that typically, three iterations seem to strike an optimal balance. In summary, it is recommended to limit the number of iterations to less than 3, which can effectively balance cost and effect.

\section{Related Work}
Recent advancements in transformer-based LLMs have not only improved the model's performance across various NLP tasks \citep{vaswani2017attention, devlin2018bert, brown2020language} but have also revealed emergent capabilities, including few-shot in-context learning, zero-shot problem solving, chain of thought reasoning, instruction following, and instruction induction \citep{cobbe2021training, wei2022chain, kojima2022large, sanh2022multitask, wei2021finetuned, ouyang2022training, honovich2022instruction}. While we share the sentiment with these works about the potential of LLMs, our focus lies in enhancing their performance through prompt editing strategies.

\paragraph{Automatic prompt engineering with training.} 
Some work improves prompts by tuning soft prompts in a differentiable manner. For instance, the work \cite{lester2021power, qin2021learning} employs soft prompts to tailor the behavior of LLMs. Similarly, efforts like those of \cite{hao2022optimizing,deng2022rlprompt,zhou2022large} delve into training auxiliary models or directly training the prompt generator \cite{hao2022optimizing,wang2022self}. While these efforts show the potential of differentiable tuning, they face limitations when LLMs are only accessed via APIs, which limits access to the model's internals. Additionally, the prompts generated from such methods often yield incoherent languages \cite{hambardzumyan2021warp}. Therefore, another type of work improves prompts via discrete manipulations using Reinforcement Learning \cite{shin2020autoprompt, zhang2023tempera, deng2022rlprompt}, which requires training a reward model. However, it is challenging to train an excellent and generalizable reward model.

\paragraph{Automatic prompt engineering without training.} 
Several works have recently explored the potential of using LLMs themselves to guide prompt optimization without training \cite{PromptProgramming, Instruction_Induction}. 
The work \cite{zhou2022large} employs the Monte Carlo sampling technique for this purpose. Similarly, the work \cite{prasad2022grips} introduced an evolutionary search approach for prompts, leveraging LLM-paraphrased and swapped segments of the original prompt.
The work \cite{chen2023teaching} focuses on refining SQL-generation prompts using LLM feedback. The work \cite{APO} considers `gradients' to guide LLMs for classification tasks. However, these methods usually grapple with ambiguous semantic orientation or a confined task-specific scope.

For automatic prompt editing, EvoPrompt \cite{EvoPrompt} uses a genetic algorithm to mutate the original prompt. PROmpting \cite{PROmpting} leverages LLMs as optimizers, where the optimization task is described in natural language. Both of them are two concurrent works. The main difference between PACE and them is that EvoPrompt does not provide feedback or reflect on execution results to LLMs, similar to PACE w/o the actor and critic, whereas PROmpting only lacks reflection, akin to PACE w/o the critic. Detailed comparison results can be found in Appendix \ref{Details of Comparison}.

In this paper, PACE refines prompts for LLMs using the actor-critic paradigm, which provides effective guidance in editing and can be applied to various tasks. 

\section{Conclusion and Discussion}

In this paper, we have proposed PACE, an innovative approach to automatic prompt editing for LLMs, drawing inspiration from the Actor-Critic paradigm in reinforcement learning. Our experiments confirm the potential of PACE in significantly enhancing the effectiveness of prompts, leading to improved LLM performance across a variety of tasks. By treating the prompt as a form of policy and conceptualizing LLMs as both actors and critics, we have presented a fresh perspective on how prompts can be optimized to better guide the output of LLMs. The remarkable improvements with PACE in prompt editing and generation underscore the value of this perspective and its potential to transform the field of prompt engineering.

\newpage

\section{Limitation}
There are several limitations to our proposed PACE as follows.

First, PACE relies on the ability of LLMs to understand prompts and instructions, which is a common limitation of all current automated prompt engineering approaches. However, with the advancement of LLM technologies, more LLMs will demonstrate such capabilities, thereby broadening the application scope of PACE.

Second, due to limitations in computational resources, we are unable to run large-scale open-source LLMs. However, we evaluated PACE on four different OpenAI LLMs. The experimental results demonstrate that PACE achieved consistent and significant performance improvements across 40 tasks on two benchmarks for all four LLMs. This to some extent validates the broad applicability and effectiveness of PACE.

Third, PACE is efficient in handling medium to low-difficulty problems, but when faced with highly complex problems, we firmly believe in the power of human-machine collaboration to unleash its greater potential, ensuring optimal results.

\bibliography{ref}

\newpage
\appendix
\onecolumn
\section{Implementation Details}\label{app:imp_details}
\begin{table}[H]
\caption{
The 24-instruction induction task proposed in the work \citet{honovich2022instruction} is described in detail. For convenience, the original table in the work \citet{honovich2022instruction} is reproduced here.}
\small
\centering
\begin{tabular}{@{}p{0.12\textwidth}@{}p{0.175\textwidth}@{}p{0.375\textwidth}p{0.300\textwidth}@{}}
\toprule
\textbf{Category}                           & \textbf{Task}           & \textbf{Instruction}                                             & \textbf{Demonstration} \\
\midrule
\textit{Spelling} & First Letter & Extract the first letter of the input word. & cat $\rightarrow$  c \\
\cmidrule{2-4}
 & Second Letter & Extract the second letter of the input word. & cat $\rightarrow$  a \\
\cmidrule{2-4}
 & List Letters & Break the input word into letters, separated by spaces. & cat $\rightarrow$  c a t\\
\cmidrule{2-4}
& Starting With & Extract the words starting with a given letter from the input sentence. & The man whose car I hit last week sued me. [m] $\rightarrow$  man, me \\
\midrule
\textit{Morpho-}

\textit{syntax} & Pluralization  & Convert the input word to its plural form.                   & cat $\rightarrow$  cats \\
\cmidrule{2-4} 
                                   & Passivization  & Write the input sentence in passive form.             &
The artist introduced the scientist. $\rightarrow$  The scientist was introduced by the artist. \\
\midrule
\textit{Syntax}                             & Negation       & Negate the input sentence.   & Time is finite $\rightarrow$  Time is not finite. \\
\midrule
\textit{Lexical} 

\textit{Semantics} & Antonyms       & Write a word that means the opposite of the input word. & won $\rightarrow$ lost \\
\cmidrule{2-4}
                                   & Synonyms       & Write a word with a similar meaning to the input word.  & alleged $\rightarrow$  supposed\\
\cmidrule{2-4}
                                   & Membership    & Write all the animals that appear in the given list.    & cat, helicopter, cook, whale, frog, lion $\rightarrow$  frog, cat, lion, whale \\
\midrule
\textit{Phonetics}                          & Rhymes         & Write a word that rhymes with the input word.           & sing $\rightarrow$  ring \\
\midrule
\textit{Knowledge}                    & Larger Animal  & Write the larger of the two given animals.              & koala, snail $\rightarrow$  koala\\
\midrule
\textit{Semantics} & Cause Selection & Find which of the two given cause and effect sentences is the cause. & Sentence 1: The soda went flat. Sentence 2: The bottle was left open. $\rightarrow$  The bottle was left open.\\
\cmidrule{2-4}
& Common

Concept & Find a common characteristic for the given objects. & guitars, pendulums, neutrinos $\rightarrow$  involve oscillations.\\
\midrule
\textit{Style} & Formality & Rephrase the sentence in formal language. & Please call once you get there $\rightarrow$  Please call upon your arrival.\\
\midrule
\textit{Numerical}         & Sum            & Sum the two given numbers.   & 22 10 $\rightarrow$  32 \\
\cmidrule{2-4}
                                   & Difference     & Subtract the second number from the first.  & 32 22 $\rightarrow$  10 \\
\cmidrule{2-4}
                                   & Number to Word & Write the number in English words.  & 26 $\rightarrow$  twenty-six \\
\midrule
\textit{Multi-}

\textit{lingual} & Translation    & Translate the word into German / Spanish / French.    & game $\rightarrow$  juego\\
\midrule
\textit{GLUE} & Sentiment 

Analysis & Determine whether a movie review is positive or negative. & The film is small in scope, yet perfectly formed. $\rightarrow$  positive \\
\cmidrule{2-4}
& Sentence 

Similarity & Rate the semantic similarity of two input sentences on a scale of 0 - definitely not to 5 - perfectly. & Sentence 1: A man is smoking. Sentence 2: A man is skating. $\rightarrow$  0 - definitely not \\
\cmidrule{2-4}
& Word in Context & Determine whether an input word has the same meaning in the two input sentences. & Sentence 1: Approach a task. Sentence 2: To approach the city. Word: approach  $\rightarrow$  not the same \\
\bottomrule
\\
\end{tabular}
\label{tab:instruct_tasks_original}
\end{table}

\newpage

\begin{table}[ht!]
\caption{A detailed description of Big-Bench Instruction Induction, a clean and tractable subset of tasks with clear human-written instructions.}
\label{table:bbii_desc}
\small
\centering
\begin{tabular}{m{2.5cm}m{5.5cm}m{5cm}}
\toprule
\textbf{Name} & \textbf{Description} & \textbf{Keywords} \\
\midrule
gender inclusive sentences german & Given a German language sentence that does not use gender-inclusive forms, transform it to gender-inclusive forms & free response, grammar, inclusion, non-English, paraphrase \\
\midrule
movie recommendation & Recommend movies similar to the given list of movies & emotional intelligence, multiple choice \\
\midrule
object counting & Questions that involve enumerating objects of different types and asking the model to count them & free response, logical reasoning \\
\midrule
operators & Given a mathematical operator definition in natural language, apply it & free response, mathematics, numerical response \\
\midrule
question selection & Given a short answer along with its context, select the most appropriate question which to the given short answer & multiple choice, paraphrase, reading comprehension, summarization \\
\midrule
ruin names & Select the humorous edit that 'ruins' the input movie or musical artist name &    emotional understanding, multiple choice \\
\midrule
snarks & Determine which of two sentences is sarcastic & emotional understanding, humor, multiple choice \\
\midrule
tense & Modify the tense of a given sentence & free response, paraphrase, syntax \\
\midrule
word sorting & Sort a list of words & algorithms, free response \\
\midrule
word unscrambling & Unscramble the given letters to form an English word & free response, implicit reasoning, tokenization \\
\bottomrule
\end{tabular}
\end{table}

\section{Templates of Actor, Critic, and Update}
\label{Prompt}
The purpose of these templates is to allow LLMs to produce corresponding responses when acting as actor, critic, and update. Note that these templates are not optimal, and we can improve these templates to get better results. 
\subsection{Actor}
Instruction: [TASK\_INSTRUCTION],\\
Input: [INPUT],\\
Output:

\subsection{Critic}
I gave you an instruction:[TASK\_INSTRUCTION]. Based on this instruction they produced the following input-prediction pairs and the corresponding ground truth:\\    
Input: [INPUT],\\
Prediction: [PREDICTION],\\
Ground Truth: [GROUNDTRUTH],\\
According to Input, Prediction, and Ground Truth, give the critical advice on how to improve the instruction: 

\subsection{Update}
I gave you an instruction:[TASK\_INSTRUCTION]. Based on the instruction they produced the following critical advices: [Critical\_Advices]. Taking these critical advices into consideration, the improved instruction was:

\newpage
\section{Details of Human-written Prompt and Performance in Instruction Induction}
\label{DII}
Each task contains multiple human-written prompts and their performances on the base model.

\subsection{Case Selection}
\begin{itemize}
    \item 0.88: Which of the following sentences is the cause?
    \item 0.8: Which of the two events is the cause?
    \item 0.6: Each input consists of two sentences, where one is the cause and the other is the outcome. Write The cause sentence.
    \item 0.52: The input is a cause and effect. Write the cause.
    \item 0.36: The input is a cause and effect, write the cause.
    \item 0.2: The input consists of two sentences. One is the cause of the other. Write the cause sentence.
    \item 0.04: Find the cause in the following cause and effect pair.
    \item 0.0: Output the sentence describing the cause (the other sentence is what happened as a result).
    \item 0.0: Output the cause (other sentence describes what happened as a result).
\end{itemize}

\subsection{Starting With}
\begin{itemize}
\item 0.72: Write a word from the following sentence that starts with the bracketed letter.
\item 0.65: Output all the tokens in the input that start with the letter in [ ].
\item 0.57: Output all tokens in the sentence that start with the letter in [ ].
\item 0.55: Write all the words of the input that start with the letter in the square brackets.
\item 0.43: Write all the words from the sentence that start with the letter in the square brackets.
\item 0.4: For each input, list all the words in the sentence that begin with the character in brackets at the end of the sentence.
\item 0.36: Write all the words in the following sentence that start with the bracketed letter, in their original order.
\item 0.35: For each input sentence, list all the words in the sentence that begin with the character written inside the brackets.
\end{itemize}

\subsection{Sum}
\begin{itemize}
    \item 1.0: You are given two numbers as input. Apply the + operator to them and output the answer.
    \item 1.0: For each input, write the sum of the two numbers that appears there.
    \item 0.7: Write the result of adding the two numbers.
    \item 0.51: Write the sum of the pair of numbers for each input.
    \item 0.29: sum the numbers in the input.
    \item 0.24: Add the following numbers.
    \item 0.19: Apply the + operator on the two numbers.
    \item 0.07: Write the sum of the two numbers.
\end{itemize}

\subsection{Rhymes}
\begin{itemize}
\item 0.62: What is a word that rhymes with the input token.
\item 0.62: Write a word that rhymes with the input.
\item 0.6: Write a word that rhymes with the input.
\item 0.6: Write a word that rhymes with the input word.
\item 0.6: Write a word that rhymes with each of the following input words.
\item 0.59: For each word in the input write another word that rhymes with it.
\item 0.58: Write a word that rhymes with the input word.
\item 0.0: Write a rhyme for the following word.
\end{itemize}

\subsection{Negation}
\begin{itemize}
    \item 0.79: Write a negated version of the given sentence.
    \item 0.79: Negate the given sentence.
    \item 0.79: Negate the following sentence:.
    \item 0.78: Write the negation.
    \item 0.72: Change the fact stated in the sentence to an opposite fact.
    \item 0.69: Output the negation of the input.
    \item 0.68: You will be given a sentence that states a fact (that might be true or not). Try to state the opposite fact.
    \item 0.5: For each input, write a sentence that expresses the exact opposite meaning of the input.
\end{itemize}

\subsection{Sentiment}
\begin{itemize}
    \item 0.91: Write "positive" if the input is a positive review, and "negative" if the input is a negative review.
    \item 0.87: Determine whether the sentiment is positive or negative.
    \item 0.87: Classify the sentiment of the input sentence (options are positive or negative).
    \item 0.85: Output whether the sentiment is positive or negative.
    \item 0.85: Given an input text, output whether the sentiment is positive or negative.
    \item 0.82: For each input, determine if the sentiment in the input is prone to negative or positive opinion.
    \item 0.76: Output whether the sentiment of the input sentence is positive or negative.
    \item 0.5: For each input, determine whether it expresses a positive or a negative opinion.
\end{itemize}

\subsection{Membership}
\begin{itemize}
    \item 0.98: Write all animals from the list of words.
    \item 0.96: Write only the animals from the list of words.
    \item 0.95: Extract animals.
    \item 0.93: List the animals from the given words.
    \item 0.91: List which of the following are animals.
    \item 0.9: Find the animals in the following list of words.
    \item 0.89: Extract all animals from the input list.
    \item 0.86: Extract all animals from the list.
    \item 0.42: Find the animals in the list.
\end{itemize}

\subsection{Large Animal}
\begin{itemize}
    \item 0.93: Write which of the pair of animals in each input is larger.
    \item 0.93: Write the bigger animal of the two.
    \item 0.93: Write the bigger animal.
    \item 0.93: For each input, write which of the two animals is bigger.
    \item 0.59: find the larger between the following pair of animals.
    \item 0.52: output which of the animals in the input is bigger.
    \item 0.46: Which is bigger?
    \item 0.4: Which of the following animals is bigger?
    \item 0.2: which of the animals separated by , is bigger.
\end{itemize}

\subsection{Word in Context}
\begin{itemize}
    \item 0.57: Each input consists of two sentences (Sentence 1 and Sentence 2), and a word that appears in at least one sentence as is or in a modified way (Word). Classify whether the meaning of this word is the same in both sentences (options are "same" or "not the same").
    \item 0.53: Write "same" if the word has the same meaning in both sentences, otherwise write "not the same".
    \item 0.52: Each input consists of two sentences (Sentence 1 and Sentence 2) and a word that appears in both of them (Word). Classify whether the meaning of this word is the same in both sentences (options are "same" or "not the same").
    \item 0.5: Given two sentences and a common word, output "same" if the common word has the same meaning in both sentences, and "not the same" otherwise.
    \item 0.49: Given two sentences and a common word, output "same" if the common word has the same meaning in both sentences, otherwise output "not the same".
    \item 0.48: "same" if the word has the same meaning in both sentences, otherwise "not the same".
    \item 0.0: Whether the meaning of the word is the same or not in both sentences.
    \item 0.0: For each input, determine whether the two sentences (marked 'Sentence 1' and 'Sentence 2') use the selected word (marked 'Word:') with the same meaning or not.
    \item 0.0: For each input, determine if the keyword (marked in 'Word:') is used in the same meaning in both the sentences (marked 'Sentence 1' and 'Sentence 2').
    \item 0.0: Determine whether the meaning of the word is the same in both sentences.
\end{itemize}

\subsection{Sentence Similarity}
\begin{itemize}
    \item 0.38: Rate from 0 (definitely not) to 5 (perfectly) the degree in which both sentences describe the same event.
    \item 0.37: Rate from 0 (definitly not) to 5 (perfectly) the degree in which the two sentences describe the same thing.
    \item 0.26: Each input consists of two sentences (Sentence 1 and Sentence 2). Rate on a scale of 0 to 5 whether Sentence 1 is a paraphrase of Sentence 2.
    \item 0.22: Score from 0 to 5 whether the two sentences describe the same event.
    \item 0.2: Score from 0 to 5 whether the two sentences describe the same event (5 being highest and 0 lowest).
    \item 0.0: Each input consists of two sentences (Sentence 1 and Sentence 2). Rate on a scale of 0 to 5 whether those sentences are paraphrases of each other, and also give a brief textual description of the rating (0 being definitely not, 2 being possibly, 3 being probably, 4 being almost perfectly and 5 being perfectly). Use " - " to separate them.
\end{itemize}

\newpage

\section{Comparison of PACE and Other Methods}
\label{Details of Comparison}
We try to reproduce these two concurrent works (i.e. PROmpting \cite{PROmpting} and EvoPrompt \cite{EvoPrompt}) following the original prompt and pseudo code in these papers.
We conduct the comparison experiment on the public benchmark - Instruction Induction, following the experimental setup in our paper. Specifically, we keep the setups consistent for all methods, including base LLM = ChatGPT (i.e., `gpt-3.5-turbo-0301'), number of input prompts = 1 (the default setups for PROmpting and EvoPrompt are 20 and 10 respectively, but many tasks do not have so many human-written prompts, so we only input the Worst Human-Written Prompt), the number of output candidate prompts = 8, and the number of iteration rounds = 1. 

The experimental results show that under the same setups, the performance improvement of PACE is significantly better than PROmpting \cite{PROmpting} and EvoPrompt \cite{EvoPrompt}, benefiting from real feedback (actors) and reflection (critics) of LLMs. In contrast, the inferior performance of PROmpting and EvoPrompt could be linked to their reliance on a substantial volume of human-written prompts. However, due to the limitations of the dataset, we only entered a single human-written prompt in this experiment.

\begin{table}[h]
\caption{Comparison between PACE versus PROmpting \cite{PROmpting} and EvoPrompt \cite{EvoPrompt}.}
\small
\centering
\begin{tabular}{lcccc}
\toprule
\textbf{Instruction}             & \textbf{Worst Human-Written Prompt} & \textbf{PROmpting} & \textbf{EvoPrompt} & \textbf{PACE} \\ 
\midrule
\textbf{active\_to\_passive}              & 1                                   & 1                      & 1                      & 0.99          \\ 
\midrule
\textbf{antonyms}                         & 0.77                                & 0.82                   & 0.82                   & 0.85          \\ 
\midrule
\textbf{cause\_and\_effect}               & 0                                   & 0                      & 0                      & 0.53          \\ 
\midrule
\textbf{common\_concept}                  & 0.05                                & 0.08                   & 0.06                   & 0.06          \\ 
\midrule
\textbf{diff}                             & 0.87                                & 1                      & 1                      & 1             \\ 
\midrule
\textbf{first\_word\_letter}              & 0.6                                 & 0                      & 0.2                    & 1             \\ 
\midrule
\textbf{informal\_to\_formal}             & 0.46                                & 0.53                   & 0.53                   & 0.59          \\ 
\midrule
\textbf{larger\_animal}                   & 0.2                                 & 0.59                   & 0.07                   & 0.53          \\ 
\midrule
\textbf{letters\_list}                    & 0.56                                & 0.48                   & 0.46                   & 1             \\ 
\midrule
\textbf{negation}                         & 0.5                                 & 0.78                   & 0.23                   & 0.76          \\ 
\midrule
\textbf{num\_to\_verbal}                  & 0.44                                & 1                      & 0.12                   & 1             \\ 
\midrule
\textbf{orthography\_starts\_with}        & 0.35                                & 0.37                   & 0.43                   & 0.44          \\ 
\midrule
\textbf{rhymes}                           & 0                                   & 0.02                   & 0                      & 0.56          \\ 
\midrule
\textbf{second\_word\_letter}             & 0.95                                & 0.45                   & 0.45                   & 1             \\ 
\midrule
\textbf{sentence\_similarity}             & 0                                   & 0.05                   & 0.05                   & 0.42          \\ 
\midrule
\textbf{sentiment}                        & 0.52                                & 0.13                   & 0.22                   & 0.91          \\ 
\midrule
\textbf{singular\_to\_plural}             & 0.99                                & 0.98                   & 0.97                   & 1             \\ 
\midrule
\textbf{sum}                              & 0.07                                & 0.99                   & 1                      & 1             \\ 
\midrule
\textbf{synonyms}                         & 0.11                                & 0.15                   & 0.14                   & 0.12          \\ 
\midrule
\textbf{taxonomy\_animal}                 & 0.42                                & 0.73                   & 0.73                   & 0.92          \\ 
\midrule
\textbf{translation\_en-de}               & 0.81                                & 0.82                   & 0.82                   & 0.84          \\ 
\midrule
\textbf{translation\_en-es}               & 0.87                                & 0.86                   & 0.83                   & 0.83          \\ 
\midrule
\textbf{translation\_en-fr}               & 0.88                                & 0.86                   & 0.87                   & 0.88          \\ 
\midrule
\textbf{word\_in\_context}                & 0                                   & 0                      & 0.01                   & 0.16          \\ 
\midrule
\textbf{Average}                 & \textbf{0.47}                       & \textbf{0.53}          & \textbf{0.46}          & \textbf{0.72} \\ 
\bottomrule
\end{tabular}
\end{table}

\newpage

\section{Efficiency Comparison}
We compare our proposed PACE to the previous prompt generation method APE, adhering to the experimental setup in our paper. As shown in the following table, we can find that the running time of PACE is acceptable and slightly lower than APE.

\begin{table}[h]
\caption{Efficiency Comparison of APE and PACE}
\centering
\begin{tabular}{lcc}
\toprule
\textbf{Instruction} & \textbf{APE Time (s)} & \textbf{PACE Time (s)} \\ \midrule
\textbf{antonyms} & 193.3841718 & 170.4483252 \\ \midrule
\textbf{cause\_and\_effect} & 165.4218265 & 146.9358413 \\ \midrule
\textbf{common\_concept} & 272.4779725 & 221.5746787 \\ \midrule
\textbf{diff} & 441.1818587 & 171.3038456 \\ \midrule
\textbf{first\_word\_letter} & 239.3755522 & 160.1741374 \\ \midrule
\textbf{informal\_to\_formal} & 220.3962668 & 131.0239611 \\ \midrule
\textbf{larger\_animal} & 301.950026 & 257.6089029 \\ \midrule
\textbf{letters\_list} & 196.5197048 & 144.7002583 \\ \midrule
\textbf{taxonomy\_animal} & 325.9682828 & 170.7813251 \\ \midrule
\textbf{negation} & 191.3055882 & 177.4928019 \\ \midrule
\textbf{num\_to\_verbal} & 193.1570891 & 187.8281341 \\ \midrule
\textbf{active\_to\_passive} & 192.6067982 & 200.9731977 \\ \midrule
\textbf{singular\_to\_plural} & 162.0425067 & 166.054487 \\ \midrule
\textbf{rhymes} & 262.3620207 & 323.8450603 \\ \midrule
\textbf{second\_word\_letter} & 223.0662 & 170.4889729 \\ \midrule
\textbf{sentence\_similarity} & 370.5423006 & 298.4649165 \\ \midrule
\textbf{sentiment} & 197.7973852 & 178.7309837 \\ \midrule
\textbf{orthography\_starts\_with} & 224.0993353 & 206.3742661 \\\midrule
\textbf{sum}                         & 279.8155023           & 172.6397824            \\ \midrule
\textbf{synonyms}                     & 320.181802            & 161.5457189            \\ \midrule
\textbf{translation\_en-de}           & 163.7373747           & 164.0702665            \\ \midrule
\textbf{translation\_en-es}           & 178.9501115           & 190.471509             \\ \midrule
\textbf{translation\_en-fr}           & 216.419857            & 186.9464092            \\ \midrule
\textbf{word\_in\_context}            & 374.4830073           & 247.1816087            \\ \midrule
\textbf{Average}             & \textbf{246.1351059}  & \textbf{191.9858079}   \\ \bottomrule
\end{tabular}
\end{table}

\newpage
\section{Details of Performance}
\begin{table}[h!]
\centering
\caption{Details of task Performance in Instruction Induction benchmark.}
\begin{tabular}{@{}lcccccc@{}}
\toprule
\multicolumn{1}{l}{\textbf{}}      & \multicolumn{1}{c}{\textbf{0}} & \multicolumn{1}{c}{\textbf{1}} & \multicolumn{1}{c}{\textbf{2}} & \multicolumn{1}{c}{\textbf{3}} & \multicolumn{1}{c}{\textbf{4}} & \multicolumn{1}{c}{\textbf{5}} \\ \midrule
\textbf{active\_to\_passive}       & 1                              & 1                              & 1                              & 1                              & 1                              & 1                              \\
\textbf{antonyms}                  & 0.81                           & 0.85                           & 0.87                           & 0.85                           & 0.88                           & 0.87                           \\
\textbf{cause\_and\_effect}        & 0.04                           & 0.53                           & 0.89                           & 0.93                           & 0.89                           & 0.85                           \\
\textbf{common\_concept}           & 0.06                           & 0.06                           & 0.15                           & 0.15                           & 0.15                           & 0.17                           \\
\textbf{diff}                      & 1                              & 1                              & 0.95                           & 1                              & 1                              & 0.99                           \\
\textbf{first\_word\_letter}       & 0.01                           & 1                              & 1                              & 1                              & 1                              & 1                              \\
\textbf{informal\_to\_formal}      & 0.53                           & 0.59                           & 0.59                           & 0.64                           & 0.53                           & 0.6                            \\
\textbf{larger\_animal}            & 0.07                           & 0.21                           & 0.64                           & 0.65                           & 0.63                           & 0.61                           \\
\textbf{letters\_list}             & 0.48                           & 1                              & 1                              & 1                              & 1                              & 1                              \\
\textbf{negation}                  & 0.26                           & 0.76                           & 0.75                           & 0.76                           & 0.75                           & 0.76                           \\
\textbf{num\_to\_verbal}           & 0.2                            & 1                              & 1                              & 1                              & 1                              & 1                              \\
\textbf{orthography\_starts\_with} & 0.37                           & 0.44                           & 0.42                           & 0.37                           & 0.29                           & 0.34                           \\
\textbf{rhymes}                    & 0                              & 0.56                           & 0.75                           & 0.82                           & 0.34                           & 0.4                            \\
\textbf{second\_word\_letter}      & 0.45                           & 1                              & 1                              & 1                              & 1                              & 0.99                           \\
\textbf{sentence\_similarity}      & 0.05                           & 0.42                           & 0.42                           & 0.41                           & 0.45                           & 0.46                           \\
\textbf{sentiment}                 & 0.13                           & 0.91                           & 0.94                           & 0.89                           & 0.87                           & 0.87                           \\
\textbf{singular\_to\_plural}      & 0.98                           & 1                              & 0.99                           & 0.99                           & 0.99                           & 0.98                           \\
\textbf{sum}                       & 0.99                           & 1                              & 1                              & 1                              & 1                              & 1                              \\
\textbf{synonyms}                  & 0.13                           & 0.12                           & 0.1                            & 0.32                           & 0.36                           & 0.46                           \\
\textbf{taxonomy\_animal}          & 0.74                           & 0.92                           & 0.96                           & 0.94                           & 0.83                           & 0.74                           \\
\textbf{translation\_en-de}        & 0.82                           & 0.84                           & 0.85                           & 0.9                            & 0.89                           & 0.86                           \\
\textbf{translation\_en-es}        & 0.87                           & 0.83                           & 0.86                           & 0.89                           & 0.95                           & 0.94                           \\
\textbf{translation\_en-fr}        & 0.87                           & 0.88                           & 0.81                           & 0.84                           & 0.78                           & 0.75                           \\
\textbf{word\_in\_context}         & 0                              & 0.16                           & 0.23                           & 0.23                           & 0.16                           & 0.16                           \\ \bottomrule
\end{tabular}
\end{table}

\begin{table}[h!]
\centering
\caption{Details of task Performance in Big-Bench benchmark.}
\begin{tabular}{@{}lcccccc@{}}
\toprule
\multicolumn{1}{l}{}                          & \multicolumn{1}{c}{\textbf{0}} & \multicolumn{1}{c}{\textbf{1}} & \multicolumn{1}{c}{\textbf{2}} & \multicolumn{1}{c}{\textbf{3}} & \multicolumn{1}{c}{\textbf{4}} & \multicolumn{1}{c}{\textbf{5}} \\ \midrule
\textbf{gender\_inclusive\_sentences\_german} & 0.175                          & 0.2                            & 0.2                            & 0.2                            & 0.225                          & 0.225                          \\
\textbf{hyperbaton}                           & 0                              & 0.14                           & 0.4                            & 0.51                           & 0.57                           & 0.51                           \\
\textbf{movie\_recommendation}                & 0                              & 0.2                            & 0.29                           & 0.27                           & 0.21                           & 0.27                           \\
\textbf{object\_counting}                     & 0                              & 0.5                            & 0.47                           & 0.44                           & 0.41                           & 0.46                           \\
\textbf{operators}                            & 0                              & 0.024                      & 0                              & 0                              & 0                              & 0                              \\
\textbf{question\_selection}                  & 0                              & 1                              & 1                              & 1                              & 0.02                           & 0.98                           \\
\textbf{ruin\_names}                          & 0                              & 0.356                       & 0.6                            & 0.3                            & 0.556                       & 0.289                       \\
\textbf{snarks}                               & 0                              & 0.514                       & 0.514                       & 0.541                       & 0.595                       & 0.568                       \\
\textbf{tense}                                & 0.728                       & 0.828                       & 0.811                       & 0.811                       & 0.811                       & 0.811                       \\
\textbf{word\_sorting}                        & 0                              & 0                              & 0.42                           & 0.46                           & 0.46                           & 0.43                           \\
\textbf{word\_unscrambling}                   & 0.19                           & 0.45                           & 0.45                           & 0.53                           & 0.58                           & 0.46                 \\ \bottomrule         
\end{tabular}
\end{table}

\end{document}